%% file: paper_main.tex
\documentclass[letterpaper, 10 pt, conference]{ieeeconf}  

\IEEEoverridecommandlockouts                              

\usepackage{times}
\usepackage[table]{xcolor}
\usepackage{soul}
\usepackage[utf8]{inputenc}

\usepackage{siunitx,  
            booktabs, 
            caption}  
            
\usepackage{comment}
\usepackage{epsfig}
\usepackage{epstopdf}
\usepackage{graphics}
\usepackage{subfigure}
\usepackage{amsmath}
\usepackage{xspace}
\usepackage{multirow}
\usepackage{hyperref}
\usepackage{float}

\usepackage{balance}

\usepackage{amssymb} 
\usepackage{pifont}
\newcommand{\cmark}{\ding{51}}%

\definecolor{lightgray}{gray}{0.85}

\newif\ifcommenton
\commentontrue
\ifcommenton
\newcommand{\red}[1]{\textcolor{red}{#1}}
\newcommand{\blue}[1]{\textcolor{blue}{#1}}
\else
\newcommand{\red}[1]{}
\newcommand{\blue}[1]{}
\fi

\renewcommand\hl[1]{#1}

\title{\LARGE \bf
MultiXNet: Multiclass Multistage Multimodal Motion Prediction

}

\author{Nemanja Djuric, Henggang Cui, Zhaoen Su, Shangxuan Wu, Huahua Wang, \\ Fang-Chieh Chou, Luisa San Martin, Song Feng, Rui Hu, Yang Xu, Alyssa Dayan, \\ Sidney Zhang, Brian C. Becker, Gregory P. Meyer, Carlos Vallespi-Gonzalez, Carl K. Wellington \\
  Uber Advanced Technologies Group\\
  \small{\texttt{\{ndjuric, hcui2, suzhaoen, shangxuan.wu, anteaglewang, fchou, luisasm\}@uber.com} }\\
  \small{\texttt{\{songf, rui.hu, yang.xu, ada, sidney, bbecker, gmeyer, cvallespi, cwellington\}@uber.com} }\\
}

\begin{document}
\maketitle
\thispagestyle{empty}
\pagestyle{empty}


\begin{abstract}
\input{abstract}
\end{abstract}

\input{intro}

\input{related}

\input{approach}

\input{eval}
\input{conclusion}




\bibliographystyle{IEEEtran}
{\footnotesize
\bibliography{references}
}

\end{document}

%% file: abstract.tex
%
%
%
%
%
One of the critical pieces of the self-driving puzzle is understanding the surroundings of a self-driving vehicle (SDV) and predicting how these surroundings will change in the near future.
To address this task we propose MultiXNet, an end-to-end approach for detection and motion prediction based directly on lidar sensor data.
This approach builds on prior work by handling multiple classes of traffic actors, adding a jointly trained second-stage trajectory refinement step, and producing a multimodal probability distribution over future actor motion that includes both multiple discrete traffic behaviors and calibrated continuous position uncertainties.
The method was evaluated on large-scale, real-world data collected by a fleet of SDVs in several cities, with the results indicating that it outperforms existing state-of-the-art approaches. 

%% file: intro.tex
\section{INTRODUCTION} 
\label{sect:introduction}

\begin{figure}[t]
    \centering
    \includegraphics[width=0.7\linewidth]{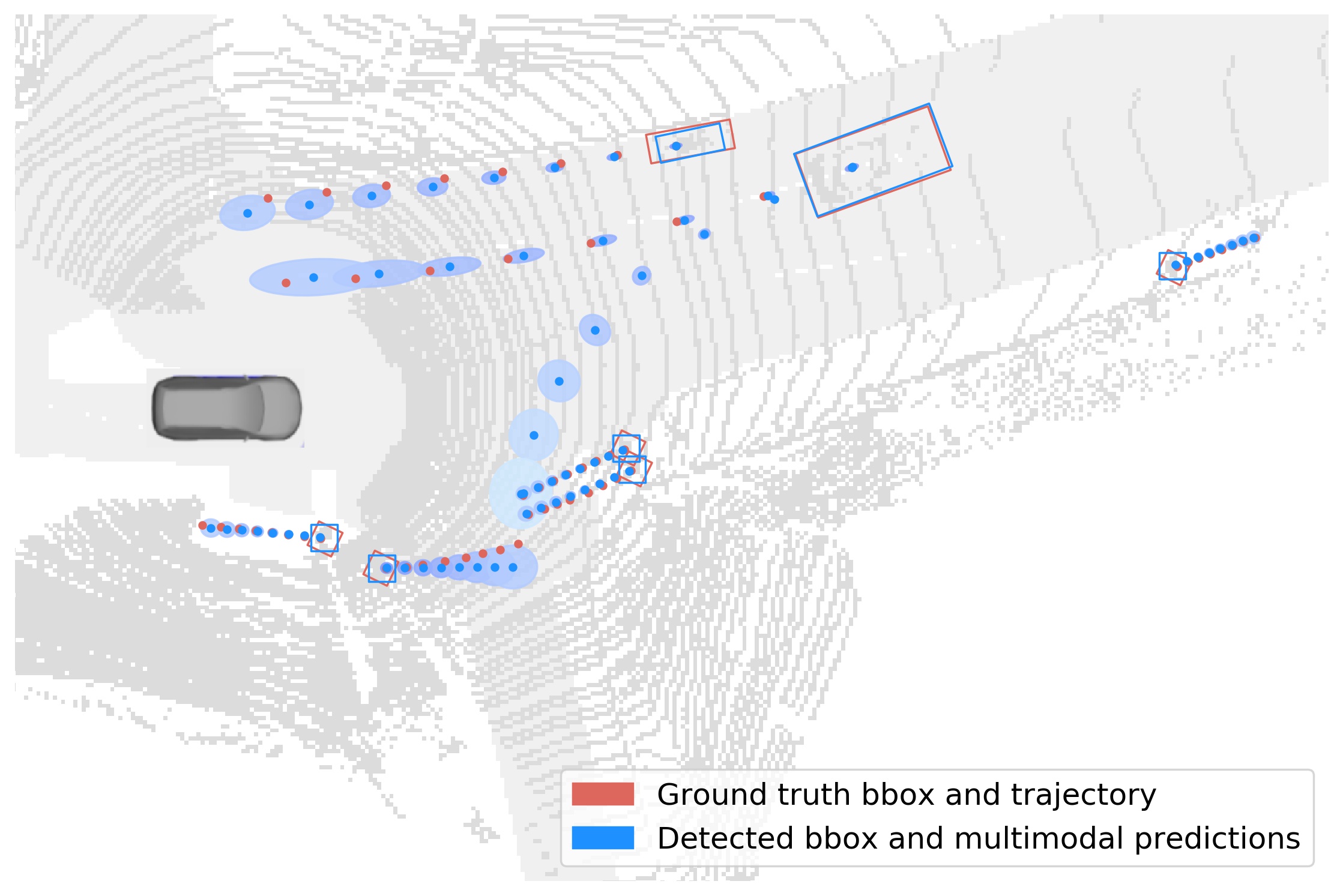}
    \caption{Example output of the proposed MultiXNet model, showing detections and multimodal, uncertainty-aware motion predictions for multiple actor types overlaid on top of lidar and map data, including pedestrians on the sidewalks and a vehicle and a bicyclist approaching the SDV; \hl{note that the turning vehicle prediction is a low-probability trajectory that is unlikely, yet still possible to observe in the real world}}
    \label{fig:bev_example}
    \vspace{-0.5cm}
\end{figure}

Predicting future states of other traffic actors, such as vehicles, pedestrians, and bicyclists, represents a key capability for the self-driving technology.
This is a challenging task, found to play an important role in accident avoidance both for human drivers \cite{stahl2014anticipation,stahl2016supporting} and for their autonomous counterparts \cite{CosgunMCHDALTA17}.
Within the context of a self-driving vehicle (SDV) it is important to capture the range of possibilities for other actors, and not just a single most likely trajectory.
Consider an opposing vehicle approaching an intersection, which may continue driving straight or turn in front of the SDV (exemplified in Fig.~\ref{fig:bev_example}). 
To ensure safety, the SDV needs to accurately reason about both of these possible modes and modulate its behavior accordingly.
In addition to the discrete modes, a downstream motion planner may react differently to a prediction depending on the position uncertainty within a predicted trajectory.
As an example, if an opposing vehicle looks like it might take a wide turn and come into the SDV’s lane, the SDV can preemptively slow down to reduce the risk. On the other hand, if the prediction shows confidence that the opposing vehicle will stay in its lane the SDV could choose to maintain its current speed.

Bringing the above requirements together, Fig.~\ref{fig:bev_example} shows an example of the task addressed by this work.
The input is a map and a sequence of lidar data which are projected into a common global coordinate frame using the SDV pose.
The output is a multimodal distribution over potential future states for the other actors in the scene.
An important challenge is that various actor types such as pedestrians and vehicles exhibit significantly different behavior, while a deployed approach needs to handle all actors present in a given scene. 

Prior work in this area has demonstrated strong performance by using end-to-end methods that jointly learn detection and motion prediction directly from sensor data~\cite{luo2018fast, casas2018intentnet}, including the addition of a jointly learned refinement stage of the network that leads to improved trajectory prediction~\cite{casas2019spatially}. 
However, these approaches have generally focused only on vehicles and produce a single trajectory rather than full motion distributions.
More recent work has shown the ability to learn a continuous distribution directly from sensor data for multiple classes~\cite{meyer2020laserflow}, but the distributions are not multimodal.
Prediction methods that operate on detections rather than the raw sensor data have shown improved performance by introducing multiclass predictions~\cite{chou2020iv}, estimates of uncertainty~\cite{chai2019multipath,djuric2020}, or incorporating multiple modes~\cite{cui2019multimodal}.
While each of these concepts has been considered individually, this work looks to unify them into a single approach which we empirically show to outperform the competing baselines.

Our work builds on IntentNet~\cite{casas2018intentnet} to produce an end-to-end motion prediction model with the following contributions:
\begin{itemize}
  \item joint detection and motion prediction of multiple actor classes: vehicles, pedestrians, and bicyclists;
  \item multimodal trajectory prediction to capture distinct potential future trajectories of traffic actors along with their corresponding continuous position uncertainties.
\end{itemize}
Using a large-scale, real-world data set, the proposed approach was shown to outperform the current state-of-the-art.

%% file: related.tex
\section{RELATED WORK}
\label{sect:related_work}


Object detection is a critical task for an SDV system, with a number of papers proposed recently in the literature.
A popular approach is using a bird's-eye view (BEV) representation, where lidar points are encoded in 3D voxels \cite{casas2018intentnet}, which has a strong benefit of being a range-invariant representation of objects. 
PointPillars \cite{lang2019pointpillars} proposed to learn the BEV encoding through a computation scheme that provides better speed while keeping accuracy high.
Range view (RV) is another popular lidar point representation that provides a compact input while preserving all sensor information. 
The authors of \cite{meyer2019lasernet} showed that RV is good at detection of both near- and long-range objects, which can further be improved by combining a camera image with RV lidar \cite{Meyer_2019_CVPR_Workshops}.
Recent work applies both BEV and RV representations \cite{chen2017multi, zhou2019end}, extracting features using separate branches of the network that are fused at a later stage. 
This fusion method preserves information for both near- and long-range objects, at the cost of a more complex and heavy network structure. 
In this work we focus on the BEV approach and discuss several ideas on how to improve the current state-of-the-art.

Movement prediction is another major topic in the SDV community. 
Typically, the prediction models take current and past detections as inputs, and then output trajectories for the next several seconds.
A common approach is to train recurrent models to process the inputs and extract learned features \cite{lee2017desire, salzmann2020trajectron++, gupta2018social, zhao2019multi, sadeghian2019sophie, kosaraju2019social}.
A number of methods have been proposed that take actor surroundings and other contextual information through BEV images as an input, and extract useful scene features using convolutional neural nets (CNNs)~\cite{luo2018fast, chou2020iv, chai2019multipath, djuric2020, cui2019multimodal, salzmann2020trajectron++}.
Interestingly, the majority of former research on trajectory prediction has focused on predicting the motion of a particular
type of road actor (e.g., vehicle or pedestrian). 
However, multiple types of traffic actors exist together on public roads, and SDVs need to accurately predict all relevant actors' motions in order to drive safely.
Moreover, different actor types have distinct motion patterns (e.g., bicyclists and pedestrians behave quite differently \cite{chou2020iv}), and it is important to model them separately.
A few recent papers tackled this challenge using recurrent methods \cite{salzmann2020trajectron++, Ma2019, chandra2019traphic}.
However, unlike in our work, an existence of a detection system was assumed and they were not trained end-to-end using raw sensor data.

Another aspect of the prediction task that is important for ensuring safe SDV operations is modeling the stochasticity of traffic behavior, either by considering multimodality of actor movement (e.g., whether they are going to turn left or right at an intersection) or position uncertainty within a single mode. 
When it comes to the multimodality of future trajectories there are two common classes of approaches.
The first is the use of generative models, either explicitly with conditional variational autoencoders \cite{lee2017desire, salzmann2020trajectron++, yuan2019diverse} or implicitly with generative adversarial networks \cite{gupta2018social, zhao2019multi, sadeghian2019sophie, kosaraju2019social, wang2020kdd}. 
Once trained, trajectories are predicted by sampling from the learned distribution at inference time.
The generative models often require the system to draw many samples to ensure good coverage in the trajectory space (e.g., as many as $20$ in \cite{gupta2018social, sadeghian2019sophie}), which may be impractical for time-critical applications.
The second category of approaches directly predicts a fixed number of trajectories along with their probabilities in a single-shot manner \cite{chai2019multipath, cui2019multimodal, phan2020covernet, cui2020deep}.
The trajectories and probabilities are jointly trained with a combination of regression and classification losses, and are much more efficient than the alternatives.
As a result, most applied work follows the one-shot approach \cite{chai2019multipath, phan2020covernet}.

In addition to multimodality, it is important to capture the uncertainty of actor motion within a trajectory mode. 
This can be achieved by explicitly modeling each trajectory as a probability distribution, for example by modeling trajectory waypoints using Gaussians \cite{chai2019multipath, djuric2020, salzmann2020trajectron++, hong2019rules}.
Following a different paradigm, some researchers have proposed non-parametric approaches \cite{jain2019discrete} to directly predict an occupancy map. While parametric approaches can easily be cast into cell occupancy space the reverse is not necessarily true, limiting the applicability of such output representations in downstream modules of the SDV system.


Instead of using independent detection and motion forecasting models, some recent work has proposed to train them jointly in an end-to-end fashion, taking raw sensor data as inputs. 
This approach was pioneered in the FaF model~\cite{luo2018fast}, while IntentNet~\cite{casas2018intentnet} further included map data as an input and proposed to predict both actor trajectories and their high-level intents. 
The authors of \cite{zeng2019end} further extended this idea to an end-to-end model that also includes motion planning. 
SpAGNN \cite{casas2019spatially} introduced a two-stage model with Rotated Region of Interest (RROI) cropping, a graph neural network module to encode actor relations, as well as modeling the uncertainty of future trajectories. 
MotionNet \cite{wu2020motionnet} used a spatial-temporal pyramid network to jointly perform detection and motion prediction for each BEV grid cell. 
LaserFlow \cite{meyer2020laserflow} proposed an end-to-end model using multi-frame RV lidar inputs, unlike the other methods using BEV representations, which can also perform prediction on multiple actor types.
Compared to our method, most of the above end-to-end methods do not consider motion prediction on diverse road actor types, and none of them addresses the multimodal nature of possible future trajectories. 
The earlier work has clearly shown the promise of end-to-end approaches, with researchers looking at various aspects to improve the prediction performance. 
In this paper we propose the first model to bring these key ideas together, and show in the experimental section the benefits over the baselines.

%% file: approach.tex
\begin{figure*}[t!]
    \centering
    \includegraphics[width=0.8\textwidth]{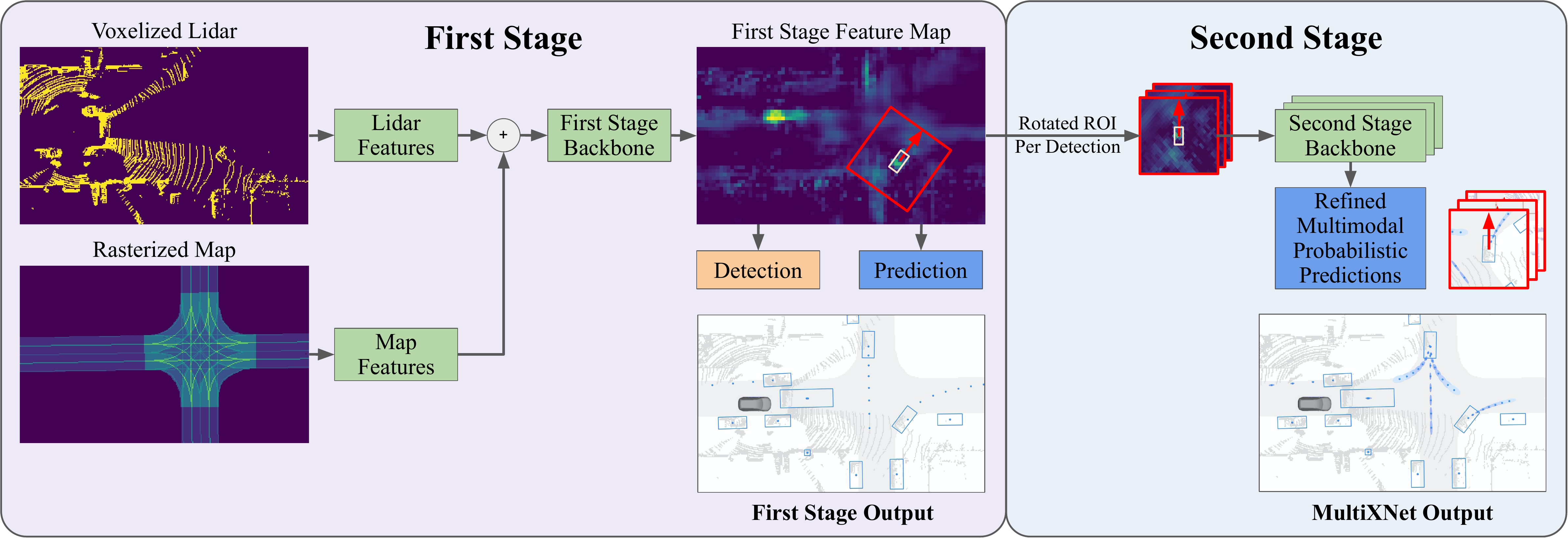}
    \caption{Overview of the MultiXNet architecture, where the first-stage network corresponding to IntentNet \cite{casas2018intentnet} outputs actor detections and their unimodal motion prediction, while the second stage refines predictions to be multimodal and uncertainty-aware; note that the first-stage prediction of the right-turning vehicle is incorrect, and the second stage improves its prediction}
    \label{fig:network_architecture}
    \vspace{-0.5cm}
\end{figure*}
    
\section{PROPOSED APPROACH}
\label{sect:methodology}

In this section we describe our end-to-end method for joint detection and prediction, called MultiXNet. 
We first describe the existing state-of-the-art IntentNet architecture~\cite{casas2018intentnet}, followed by a discussion of our proposed improvements.


\subsection{Baseline end-to-end detection and prediction}
\label{sect:e2e_intentnet}


\subsubsection{Input representation} 
In Fig.~\ref{fig:network_architecture} we show the lidar and map inputs to the network.
We assume a lidar sensor installed on the SDV that provides measurements at regular time intervals.
At time $t$ a lidar sweep $\mathcal{S}_t$ comprises a set of 3D lidar returns represented by their $(x,y,z)$ locations.
Following~\cite{casas2018intentnet}, we encode the lidar data $\mathcal{S}_t$ in a BEV image centered on the SDV, with voxel sizes of $\Delta_L$ and $\Delta_W$ along the forward $x$- and left $y$-axes, respectively, and $\Delta_V$ along the vertical axis (representing the image channels). 
Each voxel encodes binary information about whether or not there exists at least one lidar return inside that voxel.
In addition, to capture temporal information we encode the $T-1$ past lidar sweeps $\{\mathcal{S}_{t - T + 1}, \ldots, \mathcal{S}_{t - 1}\}$ into the same BEV frame using their known SDV poses, and stack them together along the channel (or vertical) dimension. 
Assuming we consider an area of length $L$, width $W$, and height $V$, this yields an input image of size $\left \lceil{\frac{L}{\Delta_L}}\right \rceil \times \left \lceil{\frac{W}{\Delta_W}}\right \rceil \times T\left \lceil{\frac{V}{\Delta_V}}\right \rceil$. 

Moreover, let us assume we have access to a high-definition map of the operating area around the SDV denoted by $\mathcal{M}$.
As shown in Fig.~\ref{fig:network_architecture}, we encode static map elements from $\mathcal{M}$ in the same BEV frame as introduced above. 
These include driving paths, crosswalks, lane and road boundaries, intersections, driveways, and parking lots, where each element is encoded as a binary mask in its own separate channel. 
This results in a total of seven map additional channels, which are processed by a few convolutional layers before being stacked with the processed lidar channels, to be used as a BEV input to the rest of the network, as described in~\cite{casas2018intentnet}.





\subsubsection{Network architecture and output}
The input BEV image can be viewed as a top-down grid representation of the SDV's surroundings, with each grid cell comprising input features encoded along the channel dimensions. 
As in~\cite{casas2018intentnet}, this image is then processed by a sequence of 2-D convolutional layers to produce a final layer that contains learned features for each cell location. 
Following an additional $1 \times 1$ convolutional layer, for each cell we predict two sets of outputs, representing object detection and its movement prediction (in the following text we denote a predicted value by the hat-notation ${\hat \ast}$).
In particular, the detection output for a cell centered at $(x, y)$ comprises an existence probability ${\hat p}$, oriented bounding box represented by its center ${\hat {\bf c}}_0 = ({\hat c}_{x0}, {\hat c}_{y0})$ relative to the center of the grid cell, size represented by length ${\hat l}$ and width ${\hat w}$, and heading ${\hat \theta}_0$ relative to the $x$-axis, parameterized as a tuple $(\sin{\hat \theta}_0, \cos{\hat \theta}_0)$.
In addition, the prediction output is composed of bounding box centers (or {\it waypoints}) ${\hat {\bf c}}_h = ({\hat c}_{xh}, {\hat c}_{yh})$ and headings ${\hat \theta}_h$ at $H$ future time horizons, with $h \in \{1, \ldots, H\}$.
A full set of $H$ waypoints is denoted as a trajectory ${\hat {\boldsymbol \tau}} = \{{\hat {\bf c}}_h, {\hat \theta}_h\}_{h=1}^H$, where the bounding box size is considered constant across the entire prediction horizon.

\subsubsection{Loss} 
\label{sect:det_loss}
As discussed in \cite{casas2018intentnet}, the loss at a certain time step consists of detection and prediction losses computed over all BEV cells.
When it comes to the per-pixel detection loss, a binary focal loss $\ell_{focal}({\hat p}) = (1-{\hat p})^\gamma \log{{\hat p}}$ is used for the probability of a ground-truth object \cite{lin2017focalloss}, where we empirically found good performance with hyper-parameter $\gamma$ set to $2$. 
Moreover, when there exists a ground-truth object in a particular cell a smooth-$\ell_1$ regression loss $\ell_1({\hat v} - v)$ is used for all bounding box parameters (i.e., center, size, and heading), where the loss is computed between the predicted value ${\hat v}$ and the corresponding ground truth $v$. 
The smooth-$\ell_1$ regression loss is also used to capture prediction errors of future bounding box centers and headings.
We refer to a cell containing an object as a {\it foreground} (fg) cell, and a {\it background} (bg) cell otherwise.
Then, the overall loss at horizon $h$ for a foreground cell $\mathcal{L}_{fg(h)}$ is computed as
\begin{align}
\begin{split}
\label{eq:loss_pred}
\mathcal{L}_{fg(h)} = ~& 1_{h=0} \Big(\ell_{focal}({\hat p}) + \ell_1({\hat l} - l) + \ell_1({\hat w} - w) \Big) + \\
& \ell_1({\hat c}_{xh} - c_{xh}) + \ell_1({\hat c}_{yh} - c_{yh}) + \\
& \ell_1(sin{\hat \theta}_h - sin{\theta}_h) + \ell_1(cos{\hat \theta}_h - cos{\theta}_h),
\end{split}
\end{align}
where $1_c$ equals $1$ if the condition $c$ holds and $0$ otherwise. 
The loss for a background cell equals $\mathcal{L}_{bg} = \ell_{focal}(1 - {\hat p})$.


Lastly, to enforce a lower error tolerance for earlier horizons we multiply the per-horizon losses by fixed weights that are gradually decreasing for future timesteps, and the per-horizon losses are aggregated to obtain the final loss,
\begin{equation}
    \mathcal{L} = 1_{\mbox{\footnotesize bg cell}} \mathcal{L}_{bg} + 1_{\mbox{\footnotesize fg cell}} \sum_{h=0}^H \lambda^{h}  \mathcal{L}_{fg(h)},
\end{equation}
where $\lambda \in (0, 1)$ is a constant decay factor (set to $0.97$ in our experiments). 
The loss contains both detection and prediction components, and all model parameters are learned jointly in an end-to-end manner.

\subsection{Improving end-to-end motion prediction}
\label{sect:improving_intentnet}
In this section we present an end-to-end method that 
we build on the approach presented in the previous section, extending it to significantly improve its prediction performance.

\subsubsection{Uncertainty-aware loss}
\label{sect:uncertain_loss}
In addition to predicting trajectories, an important task in autonomous driving is the estimation of their spatial uncertainty.
This is useful for fusion of results from multiple predictors, and is also consumed by a motion planner to improve SDV safety.
In earlier work \cite{djuric2020} it was proposed as a fine-tuning step following training of a model that only considered trajectory waypoints without uncertainties.
Then, by freezing the main prediction weights or setting a low learning rate, the uncertainty module was trained without hurting the overall prediction performance.

In this paper we describe a method that learns trajectories and uncertainties jointly, where we decompose the position uncertainty in the along-track (AT) and cross-track (CT) directions \cite{gong2004methodology}. 
In particular, a predicted waypoint ${\hat {\bf c}}_h$ is projected along AT and CT directions by considering the ground-truth heading $\theta_h$, and the errors along these directions are assumed to follow a Laplace distribution $Laplace(\mu, b)$,
where mean $\mu$ and diversity $b$ are the Laplace parameters.
We assume that AT and CT errors are independent, with each having a separate set of Laplace parameters. 
Taking AT as an example and assuming an error value ${\hat e}_{AT}$, this defines a Laplace distribution $Laplace({\hat e}_{AT}, {\hat b}_{AT})$.
Then, we minimize the loss by minimizing the Kullback–Leibler (KL) divergence between the ground-truth $Laplace(0, b_{AT})$ and the predicted $Laplace({\hat e}_{AT}, {\hat b}_{AT})$, computed as follows \cite{meyer2019alternative},
\begin{align}
\label{eg:kl_loss}
    KL_{AT} &= \log\frac{{\hat b}_{AT}}{b_{AT}} + \frac{b_{AT} \exp\left(-\frac{\left|{{\hat e}_{AT}}\right|}{b_{AT}} \right) + \left|{{\hat e}_{AT}}\right|}{{\hat b}_{AT}} - 1.
\end{align}
Similarly, $KL_{CT}$ can be computed for the CT errors, and we then use $KL_{AT}$ and $KL_{CT}$ instead of the smooth-$\ell_1$ loss for bounding box centers introduced in the previous section.
\hl{In the experimental section we compare the choice of Laplace versus Gaussian distribution, as well as the choice of KL divergence loss versus negative log-likelihood loss. The results show the benefit of optimizing Laplace KL divergence to model prediction errors.}

An important question is a choice of ground-truth diversities $b_{AT}$ and $b_{CT}$. 
In earlier detection work \cite{meyer2019lasernetkl} a percentage of label area covered by lidar points was used, however this may not be the best choice for the prediction task as the prediction difficulty and uncertainty are expected to grow with longer horizons. 
To account for this, we linearly increase the ground-truth diversity with time as
\begin{align}
\label{eq:gt_diversity}
    b_{\ast}(t) &= \alpha_{\ast} + \beta_{\ast} t,
\end{align}
where parameters $\alpha_{\ast}$ and $\beta_{\ast}$ are empirically determined, with separate parameters for AT and for CT components.
This is achieved by training models with varying $\alpha_{\ast}$ and $\beta_{\ast}$ parameters and choosing the parameter set for which the reliability diagrams \cite{djuric2020} indicate that the model outputs are the most calibrated, discussed in Sec. \ref{sect:results_quant}. 

\subsubsection{Second-stage trajectory refinement}
\label{sect:second_stage}

As shown in Fig.~\ref{fig:network_architecture}, following the detection and prediction inference described in Sec. \ref{sect:e2e_intentnet} we perform further refinement of the motion predictions for the detected objects\hl{, where we take the detected objects and infer their improved future trajectories through further processing}. 
The refinement network, which we refer to as the {\it second stage} of the model, discards the first-stage trajectory predictions and takes the inferred object center ${\hat {\bf c}}_0$ and heading ${\hat \theta}_0$, as well as the final feature layer from the main network. 
Then, it crops and rotates learned features for each actor, such that the actor is oriented pointing up in the rotated image \cite{chou2020iv,djuric2020,liang2019multi} as illustrated in Fig. \ref{fig:network_architecture}.
The RROI feature map is then fed through a lightweight CNN network before the final prediction of future trajectory and uncertainty is performed.
Both first- and second-stage networks are trained jointly, using the full loss $\mathcal{L}$ in the first stage and only the future prediction loss in the second stage, where the second-stage predictions are used as the final output trajectories.
\hl{The second stage is fully differentiable and therefore gradients flow through the second stage into the first stage.}

The proposed method has several advantages. First, the output representation can be standardized in the actor frame. 
In the first-stage model the trajectories can radiate in any direction from the actor position, while in the actor frame the majority of the future trajectories grow from the origin forward. 
In addition, the second stage can concentrate on extracting features for a single actor of interest and discard irrelevant information. 
\hl{Lastly, note that the predictions from the first and second stage can both be used in the final system, depending on the latency requirements.
For example, because the second-stage predictions incur additional latency cost, the system can use the first-stage predictions for some actors, and only run the second-stage for actors where this refinement is deemed important.}


\subsubsection{Multimodal trajectory prediction}
\label{sect:mmodal}

Traffic behavior is inherently multimodal, as traffic actors at any point may make one of several movement decisions.  
Modeling such behavior is an important task in the self-driving field, with several interesting ideas being proposed in the literature~\cite{chai2019multipath, cui2019multimodal, phan2020covernet, cui2020deep}. In this paper we address this problem, and describe an approach to output a fixed number of trajectories for each detected actor along with their probabilities.
In particular, instead of outputting a single predicted trajectory in the second stage for each detected actor, the model outputs a fixed number of $M$ trajectories. 
Let us denote trajectory modes output by the model as $\{\hat{{\boldsymbol \tau}}_m\}_{m=1}^M$ and their probabilities as $\{{\hat p}_m\}_{m=1}^M$.
First, we identify one of the $M$ modes as the \emph{ground-truth mode} $m_{gt}$, for which purpose we designed a novel direction-based policy to decide the ground-truth mode.
More specifically, we compute an angle $\Delta_{\theta} = \theta_H - \theta_0$ between the last and the current ground-truth heading, where $\Delta_{\theta} \in (-\pi, \pi]$. 
Then, we divide the range $(-\pi, \pi]$ into $M$ bins and decide $m_{gt}$ based on where $\Delta_{\theta}$ falls. 
In this way, during training each mode is specialized to be responsible for a distinct behavior (e.g., for $M=3$ we have left-turning, right-turning, and going-straight modes).
\hl{Note that there is an option to learn the modes through an Expectation-Maximization scheme, or by directly learning a mixture-density network as discussed in} \cite{cui2019multimodal}.
\hl{However, we found the proposed approach to perform best in practice.}

Given the predictions and the ground-truth trajectory, and using a similar approach as discussed in \cite{cui2019multimodal}, the multimodal trajectory loss consists of a trajectory loss of the $m_{gt}$-th trajectory mode as described in Sec. \ref{sect:uncertain_loss} and a cross-entropy loss for the trajectory mode probabilities. 
%
%
%
Lastly, we continue to use unimodal prediction loss in the first stage to improve the model training, and the multimodal trajectory loss is only applied to train the second-stage network.
\hl{Note that the accuracy of first-stage detections directly impacts the quality of second-stage predictions, as the second-stage is using the first-stage outputs as its inputs.
However, we did not observe issues due to this dependency, and the experimental results show that the second-stage refinement leads to improved prediction performance.}


\subsubsection{Handling multiple actor types}
\label{sect:handling_multi_actors}
Unlike earlier work \cite{casas2018intentnet} that mostly focused on a single traffic actor type, we model the behavior of multiple actor types simultaneously, focusing on vehicles, pedestrians, and bicyclists. 
This is done by separating three sets of outputs, one for each type, after the backbone network computes the shared BEV learned features shown in Fig.~\ref{fig:network_architecture}. \hl{Note that we found that the model achieves the best performance when each class has additional, separate processing following the learned features}.
Handling all actors using a single model and in a single pass simplifies the SDV system significantly, and helps ensure safe and effective operations.
It is important to emphasize that in the case of pedestrians and bicyclists we found that a unimodal output results in the best performance, and we do not use the multimodal loss nor the refinement stage for these traffic actors. 
Thus, in our experiments we set $M=3$ for vehicles and $M=1$ for the other actor types.
Then, the final loss of the model is a sum of per-type losses, with each per-type loss comprising the detection loss as described in Sec. \ref{sect:det_loss}, as well as the uncertainty-aware trajectory loss described in Sec. \ref{sect:uncertain_loss}, Sec. \ref{sect:second_stage}, and Sec. \ref{sect:mmodal}.

%% file: eval.tex
\section{EXPERIMENTS}
\label{sect:experiments}


\begin{table*} [ht]
\small 
\caption{\hl{Comparison on nuScenes} data using the highest-probability mode, with detection performance evaluated using average precision in \% (AP) and prediction using displacement error (DE) and cross-track error (CT) at $3s$ in centimeters; results computed on the best-matching mode (i.e., min-over-$M$) for multimodal methods shown in parentheses where available}
\label{tab:metrics_sota_nuscenes}
\centering
{
  \begin{tabular}{lccccccccc}
    & \multicolumn{3}{c}{\bf Vehicles} & \multicolumn{3}{c}{\bf Pedestrians} & \multicolumn{3}{c}{\bf Bicyclists} \\
    \cmidrule(lr){2-4} \cmidrule(lr){5-7} \cmidrule(lr){8-10}
    {\bf Method} & {\bf AP} & {\bf DE} & {\bf CT} & {\bf AP} & {\bf DE} & {\bf CT} & {\bf AP} & {\bf DE} & {\bf CT} \\
    \hline
    \rowcolor{lightgray}
    SpAGNN & - & 145.0 & - & - & - & - & - & - & - \\
    IntentNet (DE) & 61.0 & 117.4 & 37.7 & 64.4 & 83.5 & 47.3 & 30.9 & 184.6 & 73.7 \\
    \rowcolor{lightgray}
    IntentNet (AT/CT) & 60.3 & 118.3 & 37.8 & 63.4 & 83.6 & 46.8 & 31.8 & \textbf{173.0} & 70.2 \\
    MultiXNet & 60.6 & \textbf{105.0} (\textbf{104.2}) & \textbf{29.7} (\textbf{29.1}) & 66.1 & \textbf{80.1} & \textbf{43.8} & 32.6 & 203.1 & \textbf{54.8} \\
    \hline
\end{tabular}
}
\end{table*}

\subsection{Experimental setup}
Following earlier work \cite{casas2019spatially} we evaluated the proposed approach using the open-sourced nuScenes data \cite{caesar2020nuscenes} and proprietary ATG4D data.
The ATG4D data was collected by a fleet of SDVs across several cities in North America using a 64-beam, roof-mounted lidar sensor. 
It contains over $1$ million frames collected from $5{,}500$ different scenarios, each scenario being a sequence of $250$ frames captured at $10Hz$. 
The labels are precise tracks of 3D bounding boxes at a maximum range of 100 meters from the data-collecting vehicle.
Vehicles are the most common actor type in the data set, with $3.2$x fewer pedestrians and $15$x fewer bicyclists.

We set the parameters of the BEV image to $L = 150 m$, $W = 100 m$, $V = 3.2 m$, $\Delta_L = 0.16 m$, $\Delta_W = 0.16 m$, $\Delta_V = 0.2 m$, and use $T = 10$ sweeps to predict $H=30$ future states at $10Hz$ (resulting in predictions that are $3s$ long).
For the second stage, we cropped a $40m \times 40m$ region around each actor.
The models were implemented in PyTorch \cite{NEURIPS2019_9015} and trained end-to-end with 16~GPUs, a per-GPU batch size of 2, Adam optimizer \cite{kingma2014adam}, and an initial learning rate of 2e-4, training for 2 epochs completing in a day.
Note that early in training the first-stage detection output is too noisy to provide stable inputs for the second-stage refinement. 
To mitigate this issue we used the ground-truth detections for the first $2.5$k iterations when training the second-stage network. 

We compared the discussed approach to our implementation of IntentNet \cite{casas2018intentnet} which we extended to support multiple classes and tuned to obtain better results than reported in the original paper. 
In addition, using the published results we compared to the recently proposed end-to-end SpAGNN method that takes into account interactions between the traffic actors \cite{casas2019spatially}. 
We evaluated the methods using both detection and prediction metrics. 
Following earlier literature for detection metrics, we set the IoU detection matching threshold to $0.7$, $0.1$, $0.3$ for vehicles, pedestrians, and bicyclists, respectively. 
For prediction metrics we set the probability threshold to obtain a recall of $0.8$ as the operational point, as proposed in~\cite{casas2019spatially}.
In particular, we report average precision (AP) detection metric, as well as displacement error (DE) \cite{alahi2016social} and cross-track (CT) prediction error at 3 seconds. 
For the multimodal approaches we report both the min-over-$M$ metrics \cite{cui2019multimodal, lee2017desire} taking the minimal error over all modes (measuring recall) and the performance of the highest-probability mode (measuring precision).

\begin{table*} [!ht]
\normalsize
\caption{\hl{Ablation study of the proposed MultiXNet on ATG4D data}; ``Unc.'' denotes uncertainty loss from Sec. \ref{sect:uncertain_loss}, ``2nd'' denotes the refinement stage from Sec. \ref{sect:second_stage}, and ``Mm.'' denotes the multimodal loss from Sec. \ref{sect:mmodal} 
}
\label{tab:metrics_ablation}
\centering
{\small
  \begin{tabular}{cccccccccccc}
    & & & \multicolumn{3}{c}{\bf Vehicles} & \multicolumn{3}{c}{\bf Pedestrians} & \multicolumn{3}{c}{\bf Bicyclists} \\
    \cmidrule(lr){4-6} \cmidrule(lr){7-9} \cmidrule(lr){10-12}
    Unc. & 2nd & Mm. & {\bf AP} & {\bf DE} & {\bf CT} & {\bf AP} & {\bf DE} & {\bf CT} & {\bf AP} & {\bf DE} & {\bf CT} \\
    \hline
    \rowcolor{lightgray}
      &   &   & 83.9 & 90.4 & 26.0 & 88.4 & 61.8 & 32.9 & 83.2 & 51.7 & 23.5 \\
    KL-L &   &   & 84.1 & 91.9 & 22.8 & 88.2 & {\bf 57.1} & {\bf 30.4} & 84.6 & 49.9 & 21.1 \\
    \rowcolor{lightgray}
      & \cmark &   & 84.6 & \bf 82.2 & 22.2 & 88.7 & 63.2 & 33.2 & 84.3 & 51.6 & 23.8 \\
    KL-L & \cmark &   & 84.4 & 83.3 & 20.4 & 88.4 & 57.6 & 30.6 & 83.9 & 52.0 & 21.7 \\
    \rowcolor{lightgray}
      & \cmark & \cmark & 84.0 & 82.4 ({\bf 81.4}) & 22.4 (21.8) & 88.5 & 62.6 & 33.0 & 84.2 & 51.2 & 23.7  \\
    KL-L & \cmark & \cmark & 84.2 & 83.1 (82.1) & {\bf 20.2} (19.8) & 88.4 & 57.2 & 30.5 & 84.6 & {\bf 48.5} & 20.7 \\
    \hline \hline
    \rowcolor{lightgray}
    KL-G & \cmark & \cmark & 85.0 & 84.5 (83.4) & 20.6 ({\bf 19.7}) & 88.7 & 58.1 & 31.3 & 84.7 & 50.4 & {\bf 20.6} \\
    NL-L & \cmark & \cmark & 84.6 & 85.3 (84.0) & {\bf 20.2} (19.8) & 88.4 & 58.4 & 31.0 & 83.9 & 50.4 & 20.9 \\
    \rowcolor{lightgray}
    NL-G & \cmark & \cmark & 84.5 & 88.5 (87.7) & 21.1 (20.7) & 88.6 & 59.3 & 32.0 & 83.5 & 51.8 & 21.0 \\
    KL-L & no rot. & \cmark & 84.4 & 86.9 (86.0) & 21.9 (21.2) & 88.6 & 57.3 & 30.7 & 84.0 & 50.4 & 21.5 \\

    \hline
\end{tabular}
}
\vspace{-0.4cm}
\end{table*}

\subsection{Results}
\label{sect:results_quant}
The evaluation results on nuScenes data for vehicles, pedestrians, and bicyclists are summarized in Table~\ref{tab:metrics_sota_nuscenes} with best prediction results shown in bold, where we compare the proposed MultiXNet to the state-of-the-art methods SpAGNN \cite{casas2019spatially} and IntentNet \cite{casas2018intentnet}. 
Note that, in addition to the baseline IntentNet that uses displacement error (DE) in its loss, we also included a version with equally-weighted AT and CT losses instead. 
This is an extension of the baseline that uses the idea presented in Sec. \ref{sect:uncertain_loss}, which was shown to perform well in our experiments.

We can see that all methods achieved similar detection performance across the board. 
Comparing the state-of-the-art methods SpAGNN and IntentNet, the latter obtained better prediction accuracy on vehicle actors. 
The authors of SpAGNN did not provide results on other traffic actors so these results are not included in the table.
Moreover, we see that IntentNet with AT/CT losses, corresponding to the model described in Sec. \ref{sect:e2e_intentnet} that does not model the uncertainty, achieved comparable DE and CT errors as the original IntentNet with DE loss, with slightly improved results for vehicles and bicyclists. 
While the improvements are not large, this model allows for different weighting of AT and CT error components. 
This trade-off is an important feature for deployed models in autonomous driving, where prediction accuracies along these two directions may have different importance (e.g., in merging scenarios AT may be more important, while we may care more about CT in passing scenarios). 
Lastly, the proposed MultiXNet outperformed the competing methods by a significant margin on all three actor types. 
Taking only vehicles into account, we see that modeling multimodal trajectories led to improvements when considering the min-over-$M$ mode (result given in parentheses), as well as the highest-probability mode, indicating both better recall and better precision of MultiXNet, respectively.

In Table \ref{tab:metrics_ablation} we present results of an ablation study of the MultiXNet improvements performed on the larger ATG4D data set, involving the components discussed in Sec. \ref{sect:improving_intentnet}.
Note that the first row corresponds to the \mbox{\it IntentNet (AT/CT)} method, while the last row of the first part of the table corresponds to the proposed MultiXNet (KL-L denotes the KL divergence loss using Laplace distribution).
We can see that all methods had nearly the same AP, which is not a surprising result since all approaches have identical detection architectures.
Focusing on the vehicle actors for a moment, modeling uncertainty led to improvements in the CT error, which decreased by 13\%. 
Introducing the actor refinement using the second-stage network resulted in the largest improvement in the DE, leading to a drop of 11\%. 
Note that such large improvements in DE and CT may translate to significant improvements in the SDV performance.

The last three rows of the upper part of Table \ref{tab:metrics_ablation} give the performance of different variants of the second-stage model. 
Similarly to the result given previously, modeling for uncertainty led to a substantial improvement of nearly 10\% when it comes to the CT error. 
This can be explained by the fact that outliers are down-weighted due to their larger variance as shown in equation \eqref{eg:kl_loss}, and thus have less impact during training as compared to the case where the variance is not taken into account.
Furthermore, in the next two rows we evaluated the models that output multimodal trajectories. 
We can see that using the highest-probability mode to measure performance gave comparable results to a unimodal alternative.
This is due to a known limitation of such an evaluation scheme, which can not adequately capture the performance of multimodal approaches \cite{cui2019multimodal, lee2017desire}. 
For this reason in the parentheses we also report min-over-$M$, a commonly used multimodal evaluation technique in the literature, which indicated improvements in both DE and CT compared to the other baselines.

\begin{figure}[t!]
    \centering

    
    \subfigure{\label{fig:RD_CT}\includegraphics[width=0.2\textwidth]{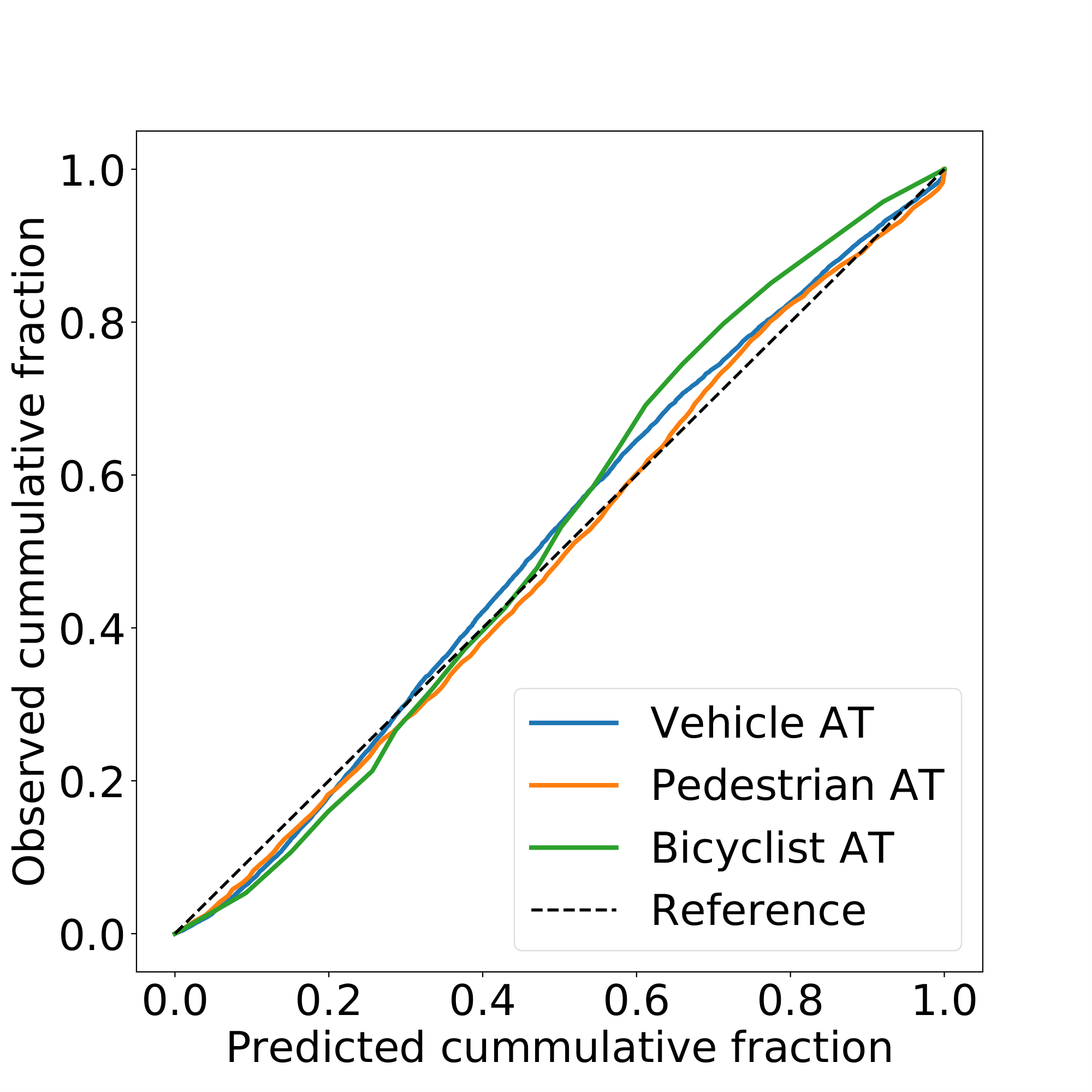}}
    \subfigure{\label{fig:RD_AT}\includegraphics[width=0.2\textwidth]{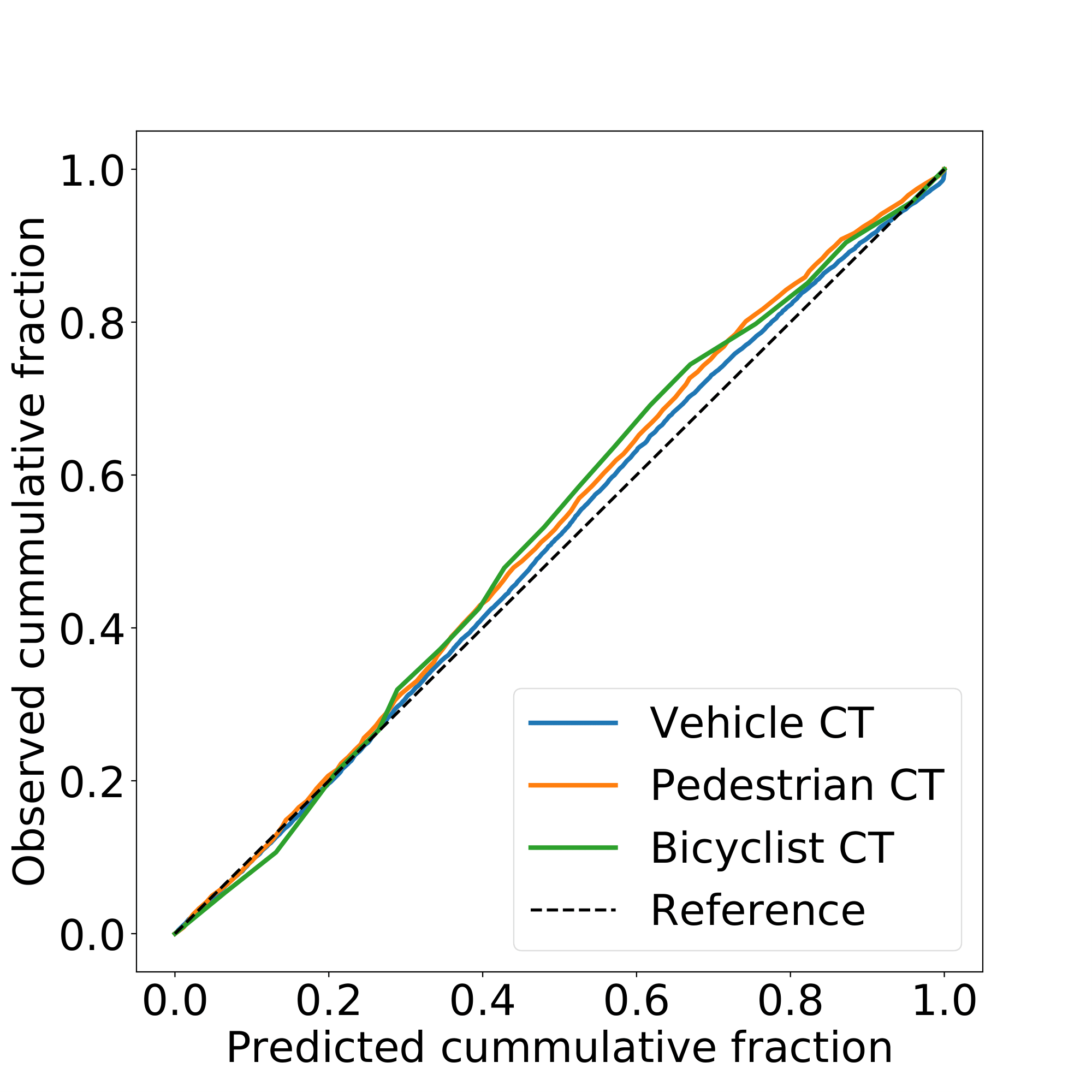}}
    
    \caption{Reliability diagrams for along-track (AT) (left) and cross-track (CT) (right) dimensions at $3s$ prediction horizon}
    \label{fig:RD}
\vspace{-0.6cm}
\end{figure}


\begin{figure*}[t!]
\centering
  \includegraphics[width=0.26\linewidth,trim={2cm 3cm 0 2.5cm},clip]{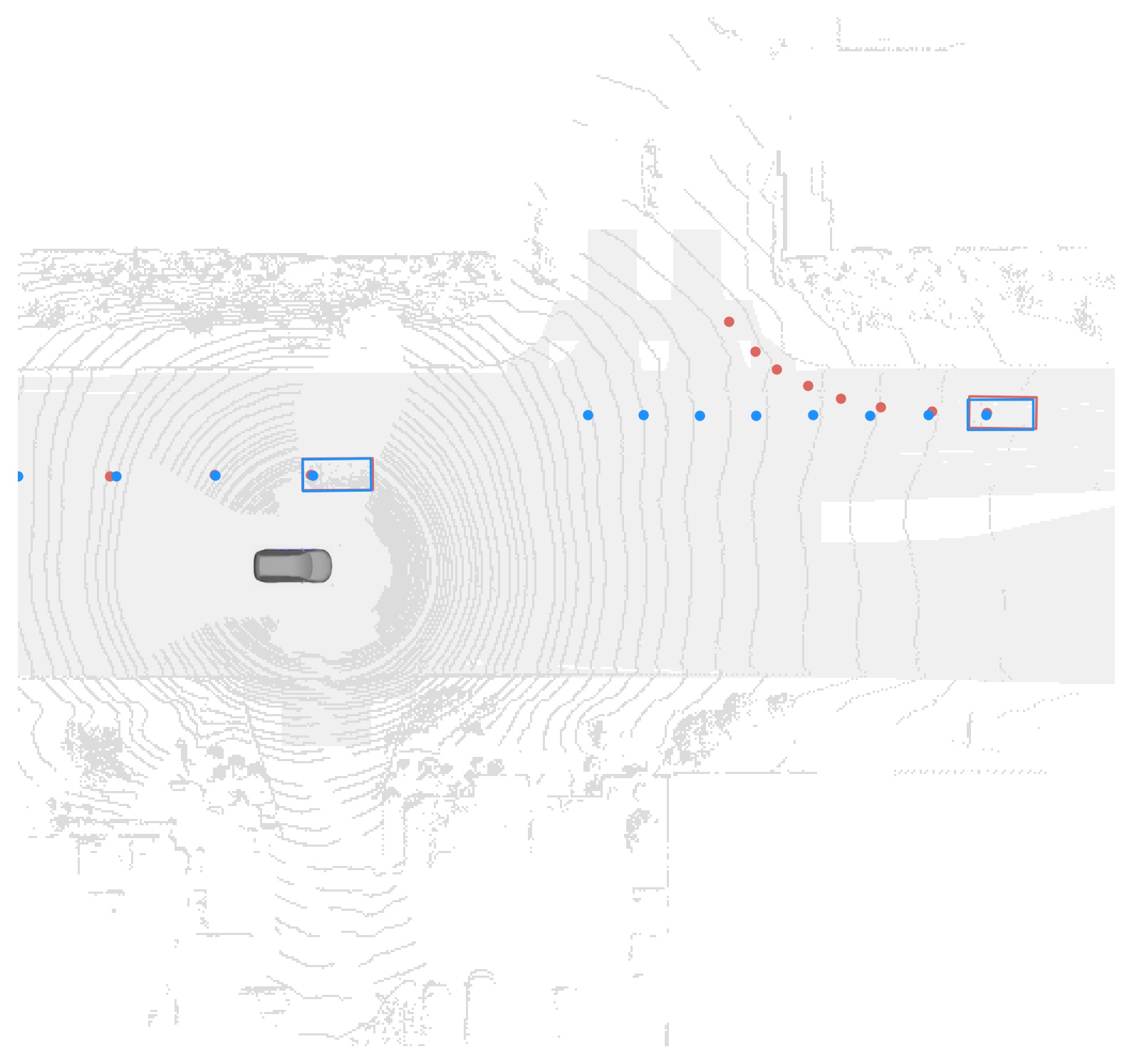} ~~~~ \includegraphics[width=0.26\linewidth,trim={4cm 4cm 0 2.0cm},clip]{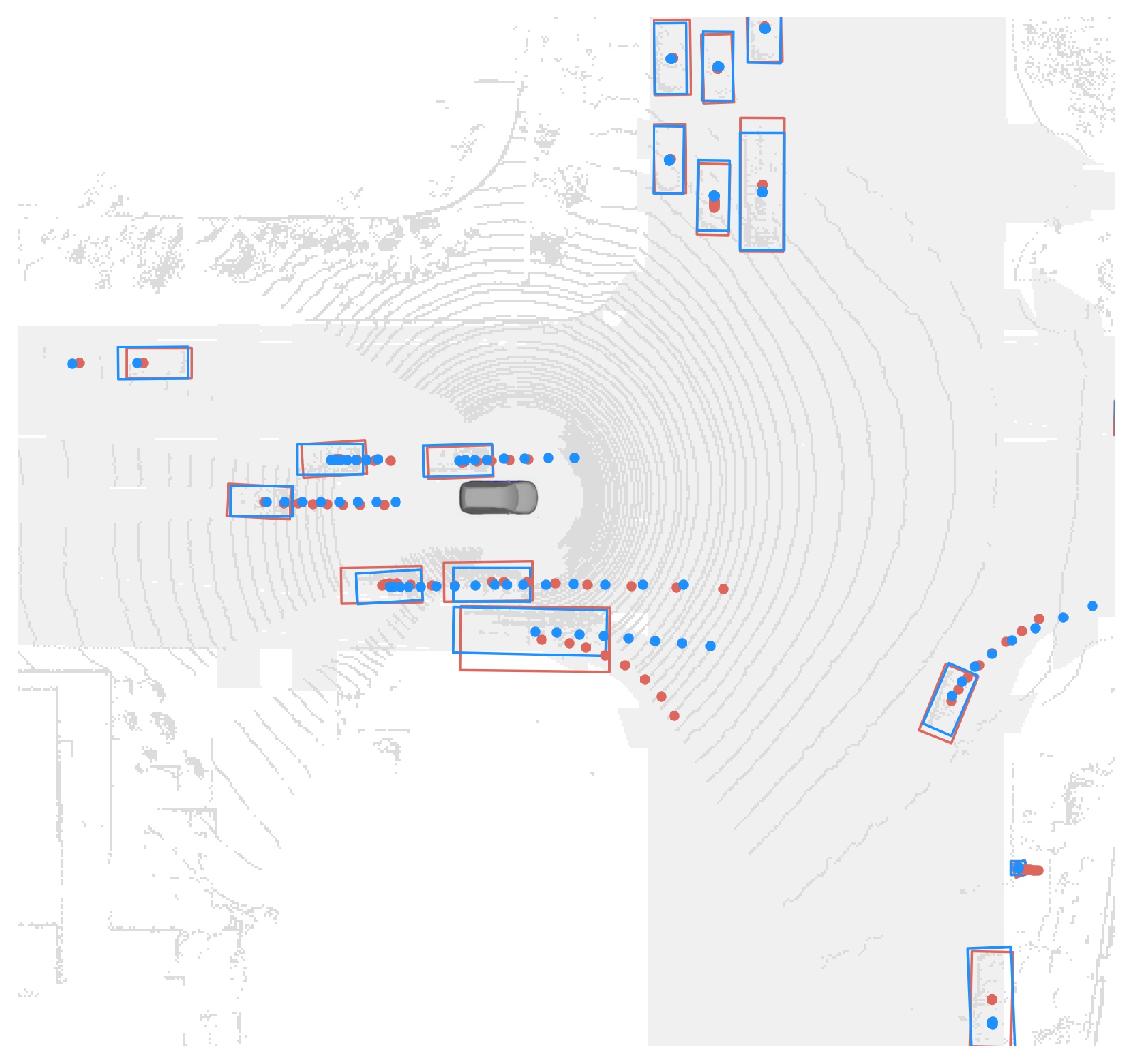} ~~~~
  \includegraphics[width=0.26\linewidth,trim={2cm 4cm 1cm 1.5cm},clip]{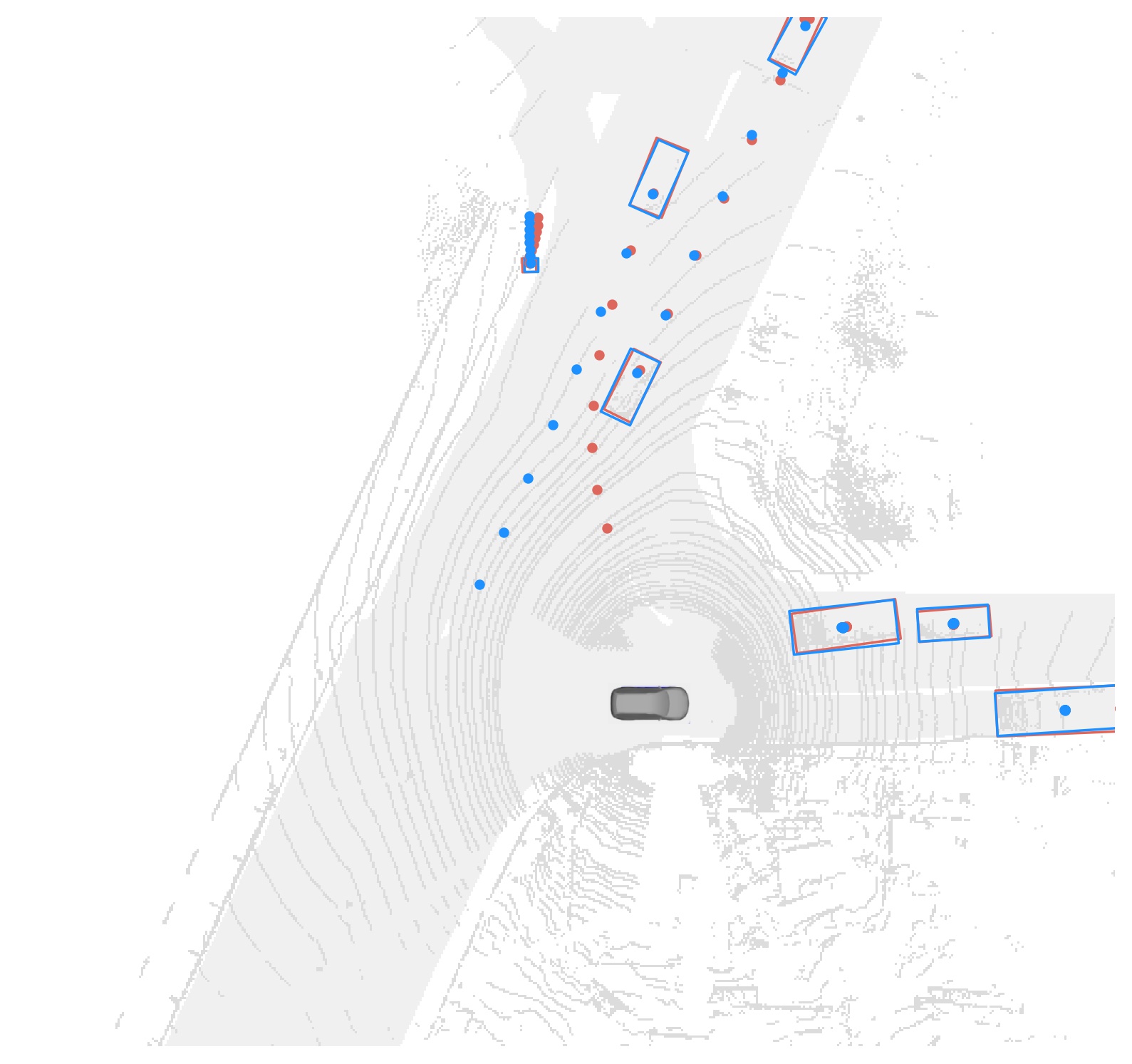} \\
  \vspace{0.1cm}
 
 \includegraphics[width=0.26\linewidth,trim={2cm 3cm 0 2.5cm},clip]{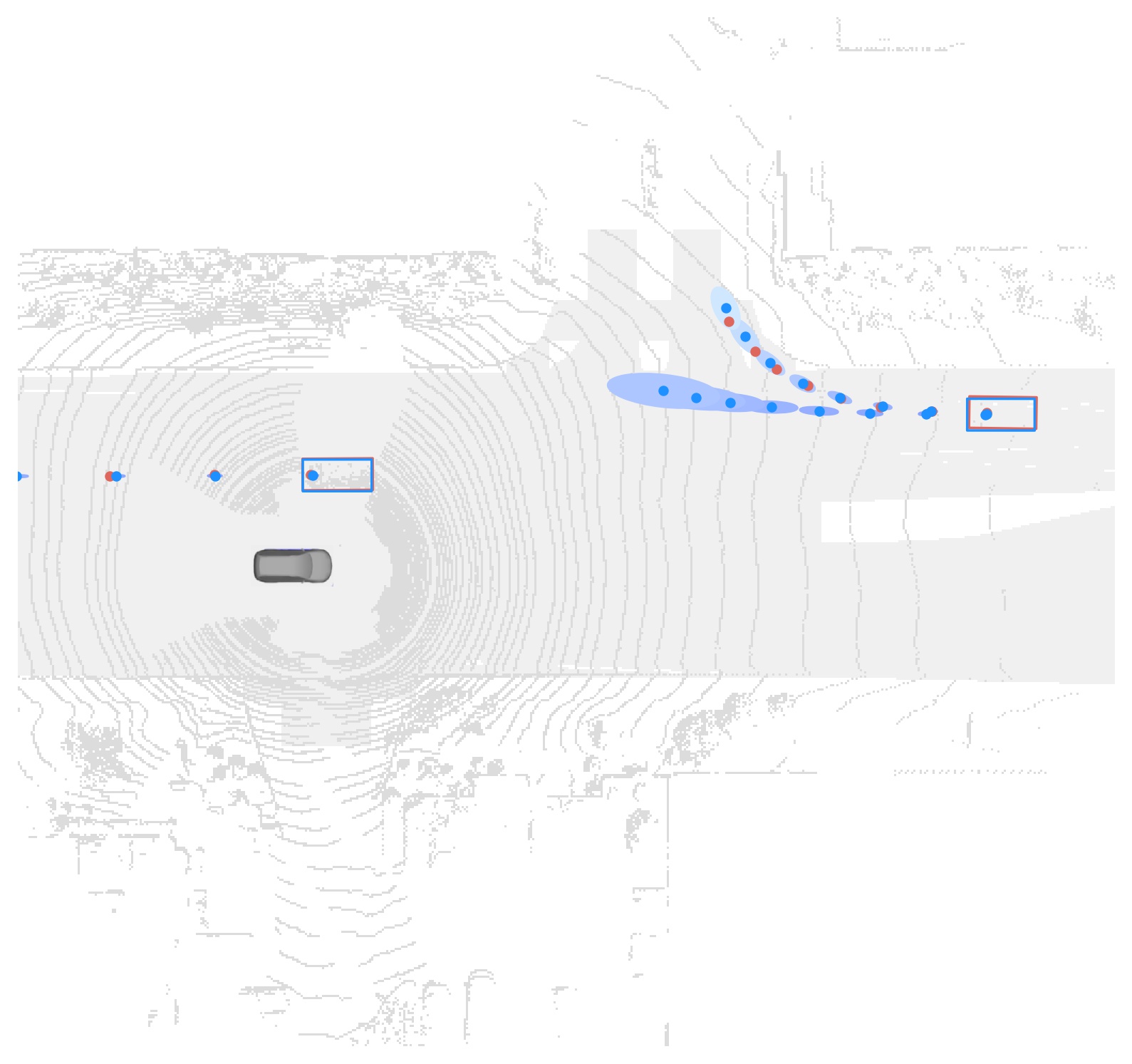} ~~~~ \includegraphics[width=0.26\linewidth,trim={4cm 4cm 0 2cm},clip]{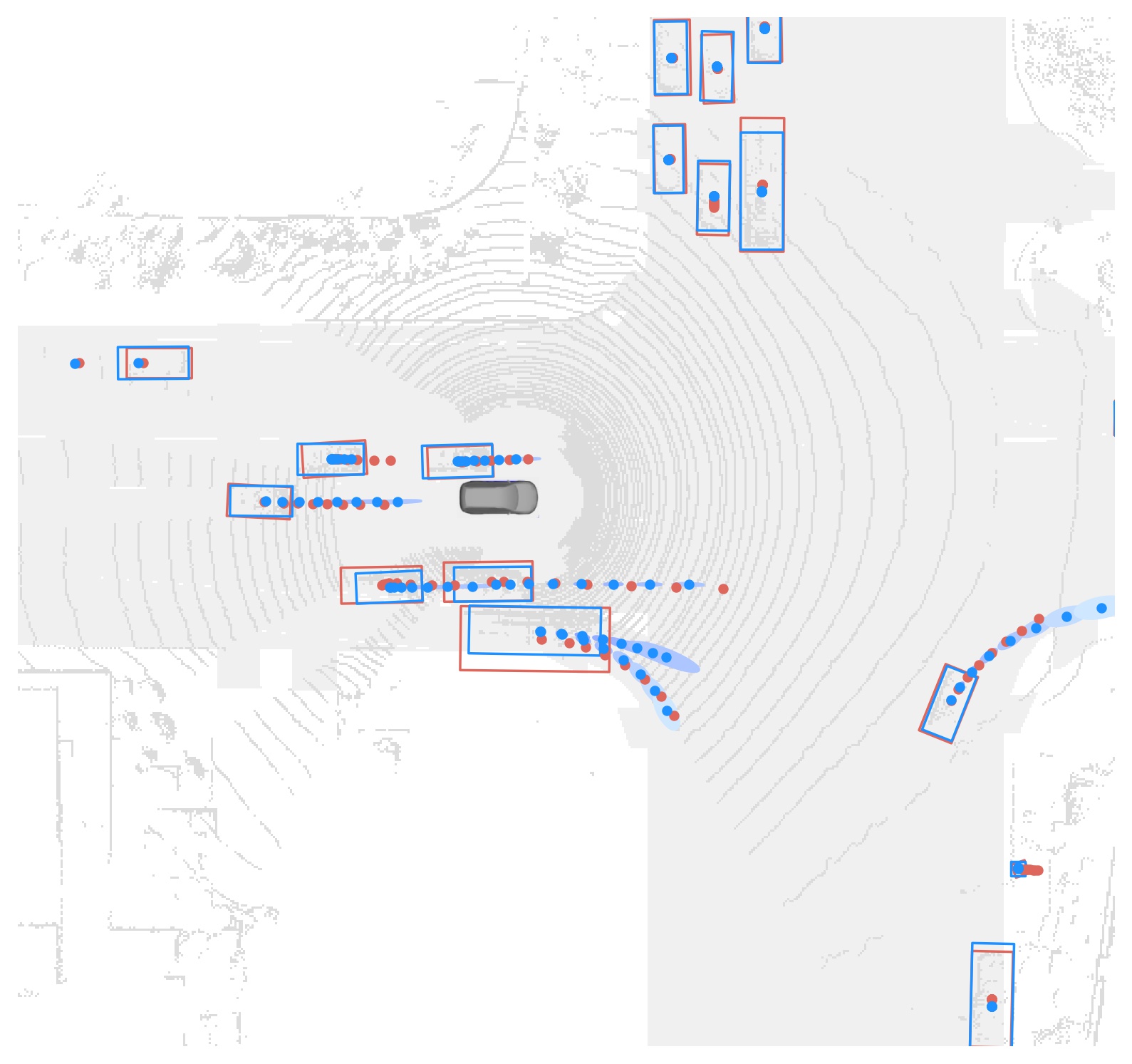} ~~~~
 \includegraphics[width=0.26\linewidth,trim={2cm 4cm 1cm 1.5cm},clip]{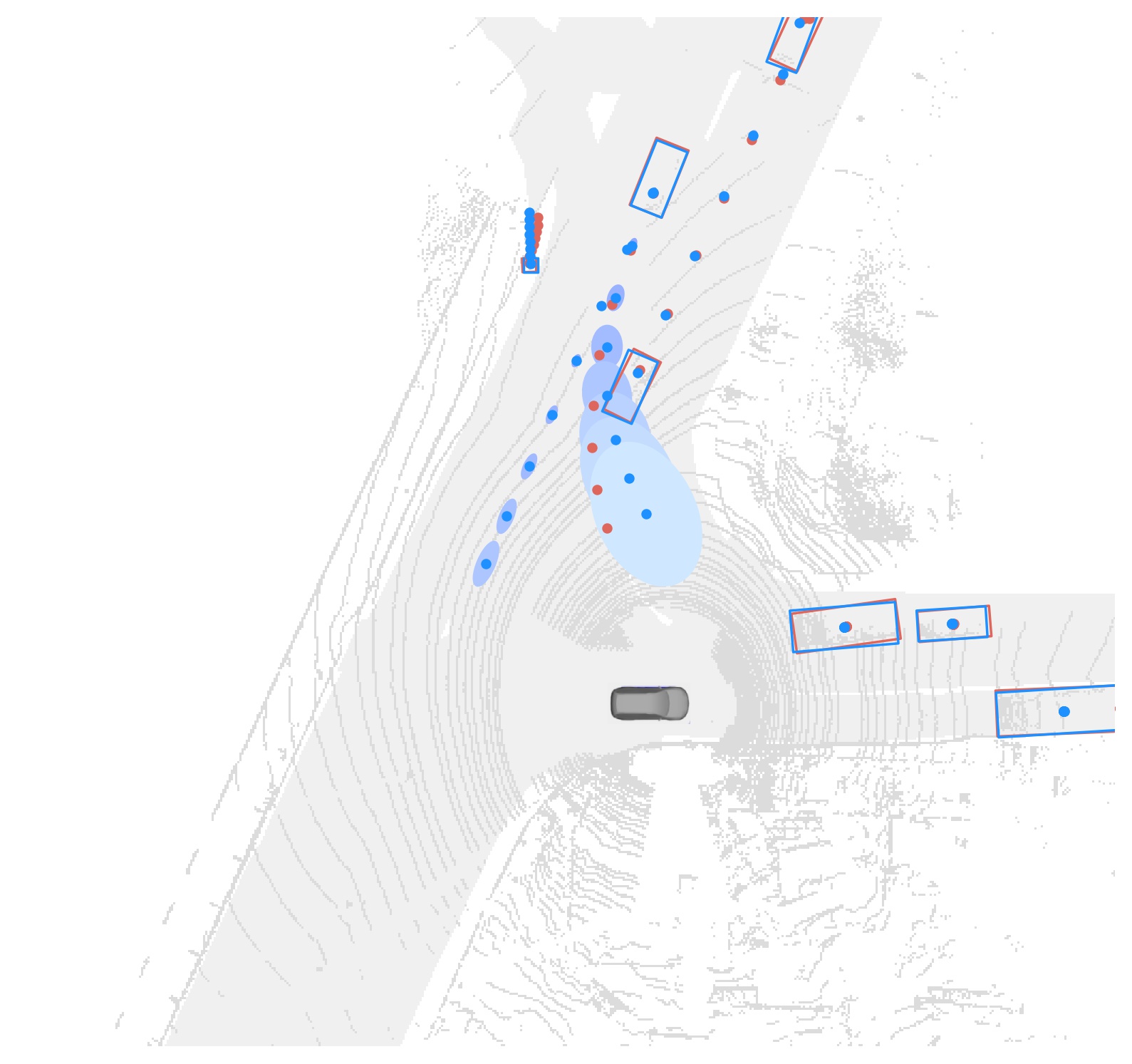} \\
 
\caption{Qualitative results of the competing models, top row: IntentNet, bottom row: MultiXNet; ground truth shown in red, predictions shown in blue, while colored ellipses indicate one standard deviation of inferred uncertainty for future predictions 
\label{fig:case_studies_all}
}
\vspace{-0.9cm}
\end{figure*}

\hl{Lastly, in the bottom four rows we perform further ablation studies by modifying the uncertainty-aware loss and the second-stage processing. 
In particular, we analyze a model using the proposed KL divergence that assumes a Gaussian noise model (KL-G) instead of Laplace (KL-L), as well as models that directly optimize negative log-likelihood with Gaussian (NL-G) and Laplace (NL-L) distributions to model the prediction errors.
We can see that using KL loss with Laplace distribution as introduced in} Sec. \ref{sect:uncertain_loss}
\hl{resulted in the lowest prediction errors across the three actor types.
Moreover, in the last row we show results of MultiXNet where we do not perform rotation of the cropped features as described in} Sec. \ref{sect:second_stage}.
\hl{We can see that the omission of rotation resulted in a significant drop in performance, further motivating the use of RROI in the second-stage processing.}

Let us discuss the results on pedestrians and bicyclists shown in the remainder of Table \ref{tab:metrics_ablation}. 
As explained in Sec. \ref{sect:handling_multi_actors}, we did not use the second-stage refinement nor the multimodal loss for these actors, and the changes indicated in the {\it 2nd} and {\it Mm.} columns only affected the vehicle branch of the network (results for the same setup changed slightly due to random weight initialization).
Similarly to the experiments with vehicles we see that modeling uncertainty led to improved results, with CT improvements between 9\% and 13\%, as seen in the second, fourth, and sixth rows.



In addition to improved performance, modeling uncertainty also allows reasoning about the inherent noise of future traffic movement. 
As mentioned in Sec.~\ref{sect:introduction}, this is an important feature that helps the motion planning generate safe and efficient SDV motions.
In Fig.~\ref{fig:RD} we provide reliability diagrams \cite{djuric2020} along the AT and CT dimensions for all three actor types, measured at the prediction horizon of 3 seconds. 
We can see that the learned uncertainties were well-calibrated, with slight under-confidence for all traffic actors. 
Bicyclist uncertainties were the least calibrated, followed by pedestrians. 
As expected, the actor types with the most training data showed the most calibrated uncertainties. 

\subsection{Qualitative results}
\label{sect:results_qual}
In this section we present several representative case studies, exemplifying the benefits of the proposed MultiXNet over the state-of-the-art IntentNet. 
Three comparisons of the two methods are shown in Fig.~\ref{fig:case_studies_all}, where we do not visualize low-probability MultiXNet trajectories below $0.3$ threshold.

In the first case, we see an actor approaching an intersection and making a right-hand turn, where unimodal IntentNet incorrectly predicted that they will continue moving straight through the intersection. 
On the other hand, MultiXNet predicted a very accurate turning trajectory with high certainty, while also allowing for the possibility of going-straight behavior. 
Apart from the predictions, we can see that both models detected the two actors in the SDV's surroundings with high accuracy.
In the second case, the SDV is moving through an intersection with a green traffic light, surrounded by vehicles. 
We can see that both models correctly detected and predicted the movements of the majority of the traffic actors. 
Let us consider motion prediction for a large truck in a right-turn lane on the SDV's right-hand side. 
Again, IntentNet predicted a straight trajectory while in actuality the actor made a turn. 
As before, MultiXNet generated multiple modes and provided reasonable uncertainty estimates for both the turning and the going-straight trajectories.

Lastly, the third case shows the SDV in an uncommon three-way intersection.
As previously, both models provided accurate detections of the surrounding actors, including one pedestrian at the top of the scene. 
Let us direct our attention to the vehicle actor approaching the intersection from the upper part of the figure. 
This actor made an unprotected left turn towards the SDV, which IntentNet mispredicted as going straight. 
Conversely, we see that MultiXNet produced both possible modes, including a turning trajectory with large uncertainty due to the unusual shape of the intersection.

%% file: conclusion.tex
\section{CONCLUSION}
\label{sect:conclusion}

We introduced MultiXNet, a multistage model that first infers object detections and predictions, and then refines these predictions using a second stage to output multiple potential future trajectories.
The proposed method was evaluated on two large-scale data sets collected on the streets of several cities, where it outperformed the existing state-of-the-art.
\hl{While obtaining good performance, there are several directions along which the method could be improved.
For example, the network does not explicitly model actor interactions, and is lidar- and map-focused while introduction of sensors such as radars and cameras could be very beneficial. 
These avenues represent areas of our future work.}